\documentclass[runningheads]{llncs}

\usepackage{graphicx}
\usepackage[ruled, linesnumbered]{algorithm2e}
\usepackage{amssymb,mathtools}
\usepackage{xcolor}
\usepackage{comment}
\usepackage{mathtools}
\usepackage[thinlines]{easytable}
\usepackage{babel,blindtext}

\usepackage{diagbox}
\usepackage{multirow}
\usepackage{longtable}
\usepackage{dblfloatfix}
\usepackage{scrextend}
\usepackage{float}
\usepackage{subcaption}
\usepackage{url}
\usepackage{amsfonts}

\usepackage{array}
\newcolumntype{L}[1]{>{\raggedright\let\newline\\\arraybackslash\hspace{0pt}}m{#1}}
\newcolumntype{C}[1]{>{\centering\let\newline\\\arraybackslash\hspace{0pt}}m{#1}}
\newcolumntype{R}[1]{>{\raggedleft\let\newline\\\arraybackslash\hspace{0pt}}m{#1}}

\begin{document}
\title{SentiLSTM: A Deep Learning Approach for Sentiment Analysis of Restaurant Reviews}
\titlerunning{Sentiment Analysis of Restaurant Reviews}
%
\author{Eftekhar Hossain\inst{1}, Omar Sharif\inst{2},
Mohammed Moshiul Hoque\inst{2*} \and Iqbal H. Sarker \inst{2}}
\institute{\textsuperscript{1}Department of Electronics and Telecommunication Engineering\\
\textsuperscript{2}Department of Computer Science and Engineering\\
Chittagong University of Engineering and Technology\\
Chittagong-4349, Bangladesh\\
\email{\{eftekhar.hossain, omar.sharif, moshiul\_240$^*$, iqbal\}@cuet.ac.bd}}
\authorrunning{Eftekhar et al.}
%

%
\maketitle              
\begin{abstract}
The amount of textual data generation has increased enor-mously due to the effortless access of the Internet and the evolution of various web 2.0 applications. These textual data productions resulted because of the people express their opinion, emotion or sentiment about any product or service in the form of tweets, Facebook post or status, blog write up, and reviews. Sentiment analysis deals with the process of computationally identifying and categorizing opinions expressed in a piece of text, especially in order to determine whether the writer’s attitude toward a particular topic is positive, negative, or neutral. The impact of customer review is significant to perceive the customer attitude towards a restaurant. Thus, the automatic detection of sentiment from reviews is advantageous for the restaurant owners, or service providers and customers to make their decisions or services more satisfactory. This paper proposes, a deep learning-based technique (i.e., BiLSTM) to classify the reviews provided by the clients of the restaurant into positive and negative polarities. A corpus consists of 8435 reviews is constructed to evaluate the proposed technique. In addition, a comparative analysis of the proposed technique with other machine learning algorithms presented. The results of the evaluation on test dataset show that BiLSTM technique produced in the highest accuracy of 91.35\%.

\end{abstract}
\section{Introduction}
Sentiment polarity detection regarded as one of the significant research problems for opinion extraction in natural language processing (NLP). In recent years, the plenteous growth of the internet and the random access of e-devices facilitate the generation of voluminous reviews or opinions in textual form on social media or online platforms. Most of these reviews express the consumers feedback toward the products and services that they received. Several business companies, as well as online marketers, take advantage of these feedbacks to provide praiseworthy services to the consumers. In addition to that customer makes a perfect decision based on the previous reviews before receiving products or services.

Sentiment detection is a computational technique that attempts to uncover the viewpoint of a user towards a specific entity. It aims to identify the contextual polarity of the text contents (such as comments, reviews, tweets or posts) as the positive, neutral and negative \cite{rahman2018aspect}. Sentiment analysis or detection has shown a remarkable impact in the business community, whereby taking into account the user opinions the communities can ensure the sustainability of their product or services. The restaurant is one such business, where customers opinions can be utilized to improve their quality of foods, environments, and services. Pompous lifestyle and assorted food habits led to a significant increase in the number of people in restaurants. To collect the excellence of services, customers instinct to look through the restaurant reviews before visit it. Therefore, reviewing a restaurant via the internet has become an ecumenical trend. Besides, an abundant amount of positive reviews can make a restaurant as a symbol of faith towards the customers. Also, it can assist a restaurant to reach the pinnacle of success. In contrast, without a sufficient amount of positive reviews, it becomes difficult to gain the attention of new customers by a restaurant. Sometimes, a restaurant with negative reviews loses the trustworthiness of the customers, which turned into reducing the profit.

Straightforwardly, users opinions on specific criteria such as food quality, ambience and service standards of a restaurant can have enough influence on the customers liking. However, it would not be wrong to say that customers inclination or reluctance towards a restaurant depends on the amount of positive and negative reviews. Therefore, the restaurants should appreciate the consents as well as the opinions of the customers. Nevertheless, scrutinizing every reviews one by one is a very time consuming as well as cumbersome task. Further, to govern such surveys, it requires plenty amount of investment in both money and human resources. Considering the fact of the explosive growth of the visitors as well as user preferences, it requires an automatic system that can comprehend the contextual polarity of reviewer opinions posted in different online platforms including Facebook, Twitter, company website, and blogs. Nevertheless, sentiment classification is a challenging research issue in a resource-poor language like Bengali. The inadequacy of benchmark dataset and the limited amount of e-textual contents or reviews in the Bengali language resulted in the sentiment classification task complicated. Deep learning algorithms are very effective to tackle such complications and classify the sentiments correctly \cite{akhtar2016hybrid,poria2020beneath}. One main advantage of these algorithms are their ability to capture the semantic information in long texts. This paper proposed a deep learning-based sentiment classification technique to classify sentiment form reviews. By taking into consideration the current constraints of sentiment analysis in low resource languages, this paper contributions illustrate in the following:
\begin{itemize}
 \item Develop a corpus consisting of $8435$ Bengali restaurant reviews which are labelled either positive or negative sentiment polarities.
 \item Develop a deep learning-based framework using LSTM for classifying the sentiment expressed in the Bengali reviews. 
   \item Perform hyperparameters tuning to settle the suitable parameters for better performance.
   \item Presents a comparative performance analysis among baseline models, including the proposed technique. 
 \end{itemize}

\section{Related Work}
Although substantial data on sentiment analysis is available in various languages, most of the research conducted on high resource languages (such as English, Chinese) \cite{bautin2008international}. Siqi Liu et al. \cite{liu2020sentiment} presented a comprehensive study of sentiment analysis on yelp restaurant reviews. They experimented different machine learning (ML) and deep learning algorithms to predict whether the sentiment of a review is positive or negative. Their work concluded that simpler models of ML more suitable than complex deep learning algorithms to predict sentiment. Minaee et al. \cite{minaee2019deep} combined  CNN and bidirectional long short term memory (BiLSTM) for sentiment analysis using IMDB movie reviews.  A cloud-based deep learning technique introduced by Ghorbani et al. \cite{ghorbani2020convlstmconv} to identify the polarity of movie reviews.  This work combined CNN and LSTM technique which achieved of 89.02\% accuracy. An integrated CNN-LSTM approach presented for sentiment analysis from the Arabic text using FastText embedding model \cite{ombabi2020deep} using. Smadi et al. \cite{al2019using} developed CNN-LSTM based technique to classify the sentiment of Arabic tweets which obtained about 64.46\% F\_1-score. A CNN with gated recurrent mechanism was introduced in \cite{xue2018aspect} to find both sentiment and aspect from restaurant review dataset. A stacked BiLSTM model proposed that incorporates continuous bag of words (CBOW) technique to determine the polarity of Chinese microblog text \cite{xu2019sentiment}. Zhou et al. \cite{zhou2019sentiment} introduced a sentiment classification model based on improved word vector representation aggregated with BiLSTM.  This work classifies the positive and negative sentiment of Chinese hotel reviews. Furthermore, Rani et al. \cite{rani2019deep} performed sentiment analysis on Hindi movie reviews using CNN which achieved 95\% accuracy for positive, negative and neutral classes.

Although sentiment analysis on resource-poor languages such as Bengali is in the preliminary stage till to date, few studies already conducted using ML and deep learning techniques. Rumman et al. \cite{chowdhury2019analyzing} presented a sentiment analysis model for Bengali movie reviews. They have experimented with different ML and deep learning algorithms on a dataset consisting of 4000 Bengali reviews and obtained accuracy of $88.90\%$ (for SVM) and $82.42\%$ (for LSTM). Eftekhar et al. \cite{hossain2020sentiment} proposed multinomial Naive Bayes model to classify sentiment polarities into the positive and negative using 2000 Bengali book reviews, which achieved $84\%$ accuracy. An LSTM-based sentiment analysis model is proposed by Hassan et al. \cite{hassan2016sentiment} for classifying Bengali text into positive and negative sentiments which achieved $78\%$ accuracy on 9337 reviews. Kamal et al. \cite{sarkar2019sentiment} proposed an LSTM based sentiment analysis to classify Bengali 1500 tweets into positive, negative, and neutral categories with an accuracy of $55.23\%$. Wahid et al. \cite{wahid2019cricket} proposed a sentiment analysis model using LSTM to classify the Bengali text into positive, negative, and neutral classes with an accuracy of $95\%$ accuracy over a dataset consists of 10,000 Facebook comments. A BiLSTM based sentiment detection method classify sentiment from Facebook tweets  \cite{sharfuddin2018deep}. This method obtained about $85\%$ accuracy on a Bengali tweet dataset consisting of 10000 tweets. Rahman et al. \cite{rahman2018aspect} introduced a sentiment analysis model for extracting positive and negative aspect category from 2000 Bengali reviews. Their system applied CNN and achieved an accuracy of $80\%$.

It is observed that most of the previous studies on sentiment analysis in the Bengali language conducted with various datasets, including Facebook tweets, book reviews, and movie reviews. However, the size of the dataset relatively was small in most recent studies. As far as we concern, there is no work except \cite{sharif2019sentiment} has explicitly done on sentiment analysis in the Bengali language that considers restaurant reviews. Though this work achieved a reasonable accuracy of $80.48\%$, there exist some notable drawbacks. Firstly, the model was trained and evaluated on a relatively tiny dataset consists of $1000$ reviews. Secondly, this work used the word frequency (TF-idf) feature extraction technique that cannot handle the disambiguation problem and thus in most of the cases the trained ML models failed to distinguish the actual sentiment of a review from an unseen data. To address these drawbacks, we proposed a deep learning-based sentiment analysis technique (Bi-LSTM) along with word2vec \cite{mikolov2013distributed} word representation algorithm  to classify the sentiment polarity into positive and negative based on our developed corpus of restaurant reviews.

\section{Corpora Development}
As Bengali is a limited resource language, it is a challenging task to accumulate a large amount of data/reviews related to the restaurant. We collect 6625 Bengali restaurant reviews from various online platforms. Among these reviews 1763, 1940 and 2922 reviews collected from restaurant pages, groups, Facebook comments respectively. We also collected 2000 restaurant reviews from the Yelp dataset \cite{asghar2016yelp}, which is manually translated into Bengali by following the method described in \cite{rahman2018aspect}. Data collection is performed between February 2020 to June 2020.

Accumulated raw data contains many inconsistent reviews. To mitigate the annotation efforts, we designed a filter to remove the inconsistencies form the collected dataset by ensuing the following steps: (i) Duplicate reviews are discarded (ii) If a review consists of less than three words, it is discarded, (iii) If a review was written in a mixture of Bengali and non-Bengali language, then it is discarded (iv) A review containing neutral sentiment is discarded, and (v) Punctuation marks, numbers, and emojis are removed from the reviews. Inconsistencies detected by the designed filter were taken care of, and a cleaned corpus is prepared for manual annotation.

After data collection and cleaning, we performed manual annotation of the 6435 reviews and did not alter the label of 2000 reviews that were collected from Yelp dataset. For annotation, we assigned three annotators having 12-18 months experience in NLP domain. For the purpose of annotation, we follow the techniques presented by Mohammad \cite{mohammad2016practical}. Annotators are asked to label a review either positive or negative. To clear the understanding of the annotators, we provide some restaurant reviews to them prior labelling. In order to check the quality of the annotation, we measure the inter-rater agreement between the annotators using Cohen's Kappa \cite{cohen1960coefficient}. The average Kappa score is of 0.81 in our corpus, which indicates that data are of acceptable quality. Finally, the label of a review is decided based on majority voting among the annotators. We developed the Bengali Restaurant Review Corpus (BRRC) containing 4865 positive and 3570 negative reviews. Some statistics of the BRRC summarized in Table~\ref{data}. 

\begin{table}[h!]
\begin{center}
\caption{Summary of the BRRC.}
\begin{tabular}{|C{4cm} | C{2cm}| C{2cm}|}
\hline
   Dataset attributes & Positive & Negative  \\
\hline
     Number of documents & 4 865 & 3 570 \\
\hline
    Number of words & 85 956 & 270 613\\
\hline
    Total unique words & 11 295 & 5 470\\
\hline
 Size (in bytes) & 8 974 848 & 579 872  \\
 \hline 
 Number of sentences & 15 081 & 40 710  \\
 \hline 
\end{tabular}
\label{data}
\end{center}
\end{table}

\section{Methodology}
\begin{figure}[h!]
\centering
  \includegraphics[height=7.5cm,width=12cm]{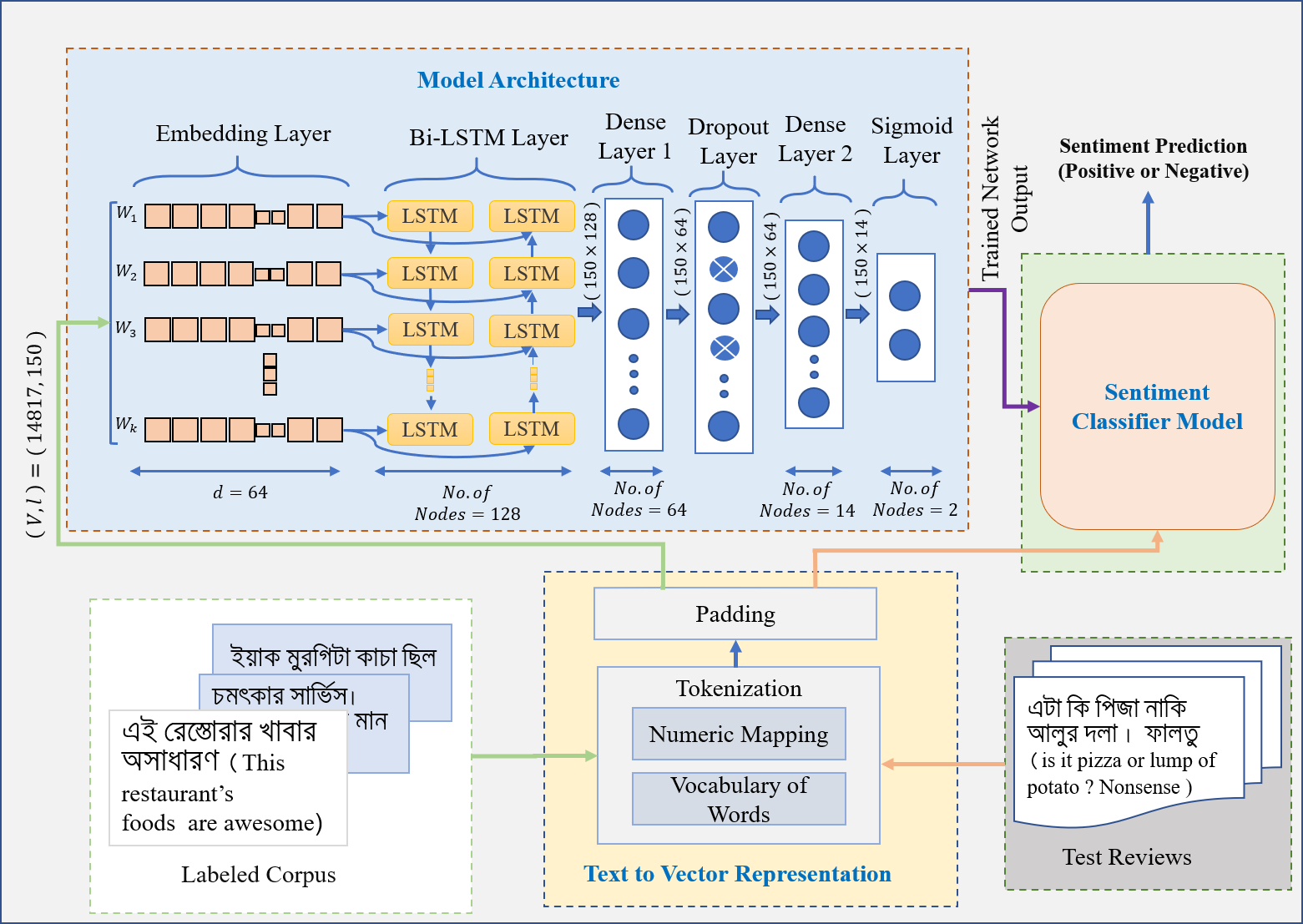}
  \caption{BiLSTM based sentiment classification framework.}
  \label{fig:proposed_model}
\end{figure}

The main objective of our work is to develop a sentiment classifier using deep learning approach (i.e Bi-LSTM) that can classify restaurant reviews into positive or negative sentiment categories. An abstract view of the proposed framework is depicted in Fig.~\ref{fig:proposed_model}. The proposed framework consists of three main constituents: text to vector representation module, model architecture module, and sentiment prediction module.

\captionsetup[figure]{labelfont={bf},labelformat={default},labelsep=period,name={Fig.}}

\subsection{Text to Vector Representation}
Success of any deep learning algorithm heavily depends on the features applied during training. As deep learning algorithms are unable to learn from raw reviews, we have to create a numeric mapping of the reviews, $R[]=\{r_1, r_2, ..., r_m\}$. To get this numeric mapping, a vocabulary $V$ of $k$ unique words is created, $V=\{u_1, u_2, ..., u_k\}$. Words $(w_i)$ in a review $r_j=[w_1, w_2, ..., w_{l^{'}}]$ replaced by the index value $(i)$ of the words in $V$. Thus we get converted vector sequence ($s^{'}$) from a review ($r_j$), $s^{'}= [i_{1}, i_{2}, ..., i_{l^{'}}]$. At this stage we get variable length sequences, $S^{'}=\{s^{'}_1, s^{'}_2, ..., s^{'}_m$ \} which are not suitable for feature extraction and training. By applying pad sequence method $(S^{'})$ transformed into fixed length sequences, $S = \{s_1, s_2, ..., s_m\}$. Each sequence ($s_k$) of $S$ is a fixed length vector of size $l$. To reduce computational cost, optimal sequence length $l$ is chosen by analyzing the length distribution of the reviews.
We observe that most of the reviews are shorter than $150$. Hence, $l = 150$ is chosen as optimal review length to keep the useful information intact as well as develop the system with minimal computation. Extra words discarded form long reviews and a zero vector padded with short reviews to maintain the length $l$.

\subsection{Model Architecture}
The model architecture comprises three major blocks: embedding layer, Bi-LSTM layer and classification layer. 

\subsubsection{Embedding Layer}
To extract the feature, we used word2vec \cite{mikolov2013distributed} embedding technique that is a way of mapping integer indices of textual data into the dense vector. We used entire BRRC to train word2vec using Keras embedding layer. Embedding layer takes three inputs $(V_k,d,l)$ where $V_k$= size of vocabulary, $d=$ embedding dimension, $l=$ length of a review. Embedding dimension $d$ is a hyper-parameter which determines the length of the vector representation of a word. Embedding layer converted a review into a 2D vector of dimension $l\times d$. Thus, for the $R$ number of reviews, we obtained a feature vector of dimension $F = R\times l\times d$.

\subsubsection{LSTM Layer}
Long short-term memory (LSTM) network is a commonly used variant of recurrent neural network (RNN) which used as a solution of exploding and vanishing gradient problem. Particularly, LSTM are proved \cite{yenter2017deep} effective to capture the long term dependencies in a text. We applied bidirectional LSTM (BiLSTM) to keep the contextual information form previous as well as next word \cite{kalchbrenner2015grid,graves2013generating}. Word embedding values of the embedding layer passed to each of the LSTM where each LSTM consists hidden units of size $h$. For BiLSTM, after the concatenation of each LSTM output we obtained a vector representation of length $2h$. LSTM process a input sequence of embedding vector as a pair $(e^{<i>},y^{<i>})$. For each pair $(e^{<i>},y^{<i>})$ and each time step $t$, a hidden vector $h^{<t>}$ and a remember vector $m^{<t>}$ is preserved by a LSTM. This vectors are responsible for regulating the updates and outputs of states. This helps to produce target output $y^{<i>}$ based on the past states of the $x^{<i>}$ input. The processing steps at time $t$ executed by the Eqs. \ref{eq:1}-\ref{eq:6}.
\begin{equation} \label{eq:1}
    u_g = \sigma(W_u * h^{<t-1>} + I_u)
\end{equation}
\begin{equation}
    f_g = \sigma(W_f * h^{<t-1>} + I_f) 
\end{equation}
\begin{equation}
    o_g = \sigma(W_o * h^{<t-1>} + I_o) 
\end{equation}
\begin{equation}
    c_g = \tanh (W_c * h^{<t-1>} + I_c) 
\end{equation}
\begin{equation}
   m^{<t>} = f_g \odot m^{<t-1>} + u_g \odot c_g 
\end{equation}
\begin{equation}\label{eq:6}
   h^{<t>}= tanh (o_g \odot m^{<t>} )
\end{equation}

Here, $\sigma$ represents the sigmoid activation function, correspondingly $W_u, W_f$, $W_o, W_c$ and $I_u, I_f, I_o, I_c$ are weight matrices and projection matrices of the recurrent units. 
The computed gates $u_g, f_g, o_g, c_g$ of LSTM cells play pivotal role in attaining significant attributes from the computed vector by storing in the remember vector $m^{<i>}$ as long as needed. The forget gate $f_g$ decides the amount of information to be dumped from the previous remember vector $m^{<i-1>}$, on the contrary the update gate $c_g$ use input gate $u_g$ and previous remember vector $m^{<i-1>}$ to write updated information in the new remember vector $m^{<i>}$. Finally, output gate $o_g$ monitors which information goes from new memory vector $m^{<i>}$ to the hidden vector $h^{<i>}$.

\subsubsection{Classification Layer}
This step includes an two dense layer followed by a  sigmoid layer. For the $r^{th}$ input sequence, BiLSTM layer takes input of size $(l \times d)$ and transformed into output vector of size $ (l \times2h )$. This vector is propagated through the first dense layer and generates a new vector of shape $(l \times dl_{1})$ with the help of rectified linear unit (Relu) activation function. To avoid over-fitting a dropout layer introduced between two dense layer with dropout ratio of 26\% \cite{gulli2017deep}. At every iteration 74\% neurons are randomly chosen to pass their output form first dense layer to second dense layer which further generates a new vector of size $(l \times dl_2)$. Noted that the $dl_1$ and $dl_2$ correspondingly represents number of hidden neurons in first and second dense layer. By applying flattening to output vector of the second dense layer we got an one dimensional vector of size $f_v$. Finally, the last layer output vector passed into a sigmoid \cite{gulli2017deep} layer. The sentiment of the  $r^{th}$ input sequence of $m$ reviews is calculated by using the equation \ref{eqn:sigm} and \ref{eqn:y_pred}.
\begin{equation} 
\label{eqn:sigm}
\sigma(f_v) = \frac{1}{1+e^{-f_v}}    
\end{equation}

\begin{equation} 
\label{eqn:y_pred}
y_{pred} = 
\begin{cases}
     1\ (positive) & \ if \ \sigma(f_v) > Threshold \\
      0\ (negative) & \ if\ \sigma(f_v) < Threshold
\end{cases}
\end{equation}

The equation \ref{eqn:loss} corresponds to the cross entropy loss function \cite{gulli2017deep} that we used to train the model. Here, $r$ subscript indicates the $r^{th}$ input review and $t_r$ indicates the true sentiment class of  $r^{th}$ review.

\begin{equation} \label{eqn:loss}
 Loss(y,y_{pred}) = -\frac{1}{R} \sum_{r=1}^{R} (t_r \ \log(y_{pred}))  
\end{equation}

\subsection{Sentiment Classifier Model}
The purpose of this module is to determine the sentiment of reviews that the model never seen before. For classification at first, an unlabeled review is feed into the text to vector representation module, where it is gone through the tokenization and padding steps. Then, the trained sentiment classifier model takes this transformed vector as an input and predicts the sentiment of that review.     

\section{Experiments}
The goal of this experiment is to find out the appropriate hyperparameter combination as well as analyze the effectiveness of the proposed model over other machine learning algorithms. We used Google Colaboratory to conduct experiments which widely used for building deep learning applications. Pandas == 1.0.5 data frame used to prepare data. Deep learning model developed on Keras == 2.3.0 and TensorFlow == 2.2.0 framework. 
Training, validation, and testing sets are consist of 72\% (6072 reviews), 18\% (1519 reviews), 10\% (844 reviews) of total reviews, respectively. Training set used to train the model while validation samples help to tune the hyper-parameters (i.e. learning rate, batch size) of the model. Finally, the trained model evaluated with the test set.

\subsection{Hyperparameter Optimization}
Model hyperparameters are the parameters that directly governs the training process of a model. These parameters determine the network architecture (i.e. number of layers, number of hidden units) and how the network is trained (i.e. batch size, learning rate). Two hidden layers respectively with $64$ and $14$ hidden units have used. The hyperparameter settings for the proposed model such as embedding dimension, batch size, dropout rate, optimizer, learning rate and the number of epochs are specified in Table~\ref{hyper}. We arbitrarily choose an initial value for each hyperparameter except embedding dimension. To find the optimum value of a hyperparameter, we iterate through the hyperparameter space. 
The proposed model is trained with these optimum hyperparameter settings.
\vspace{-0.5cm}
\begin{table}[h!]
\begin{center}
\caption{Hyperparameter settings}
\begin{tabular}{|C{3cm} | C{1.5cm}| C{5cm}|C{2cm}|}
\hline
  Hyperparameters & Initial value & Hyperparameter space & Optimal value \\
\hline
     Embedding Dimension & - & $8$, $16$, $32$, $64$, $100$, $128$, $200$, $256$, $400$, $512$, $600$, $700$, $800$, $1024$ & $128$ \\
\hline
    Batch Size & $32$ & $4$,$8$,$16$, $32$, $64$, $128$, $256$, $512$ & $64$ \\
\hline
    Dropout & $0.1$ & $0.1$, $0.15$, $0.2$, $0.23$, $0.27$, $0.3$, $0.33$, $0.36$, $0.4$, $0.43$, $0.46$, $0.5$, $0.54$, $0.57$, $0.6$, $0.63$, $0.66$, $0.69$, $0.72$, $0.75$ & $0.46$\\
\hline
 Optimizer & $Adam$ &$SGD, RMSprop, Adam, Nadam $ & $RMSprop$ \\
 \hline 
 Learning Rate & $0.01$ & $0.9$, $0.6$, $0.3$, $0.1$, $0.09$, $0.06$, $0.03$, $0.01$, $0.009$, $0.006$, $0.003$, $0.001$, $0.0009$,
            $0.0006$, $0.0003$, $0.0001$, $0.00001$, $0.000001$ & $0.0001$ \\
  \hline 
 Number of Epochs & $20$ & $4$, $6$, $8$, $10$, $12$ ,$14$ ,$16$, $18$, $20$, $25$, $30$, $35$, $40$, $45$, $50$ & $10$ \\
 \hline 
\end{tabular}
\label{hyper}
\end{center}
\end{table}
\vspace{-1.2cm}
\subsection{Results}
\begin{figure}[!b]
       \centering
\begin{subfigure}[h!]{0.49\textwidth}
\includegraphics[height =4.5cm,width=\textwidth]{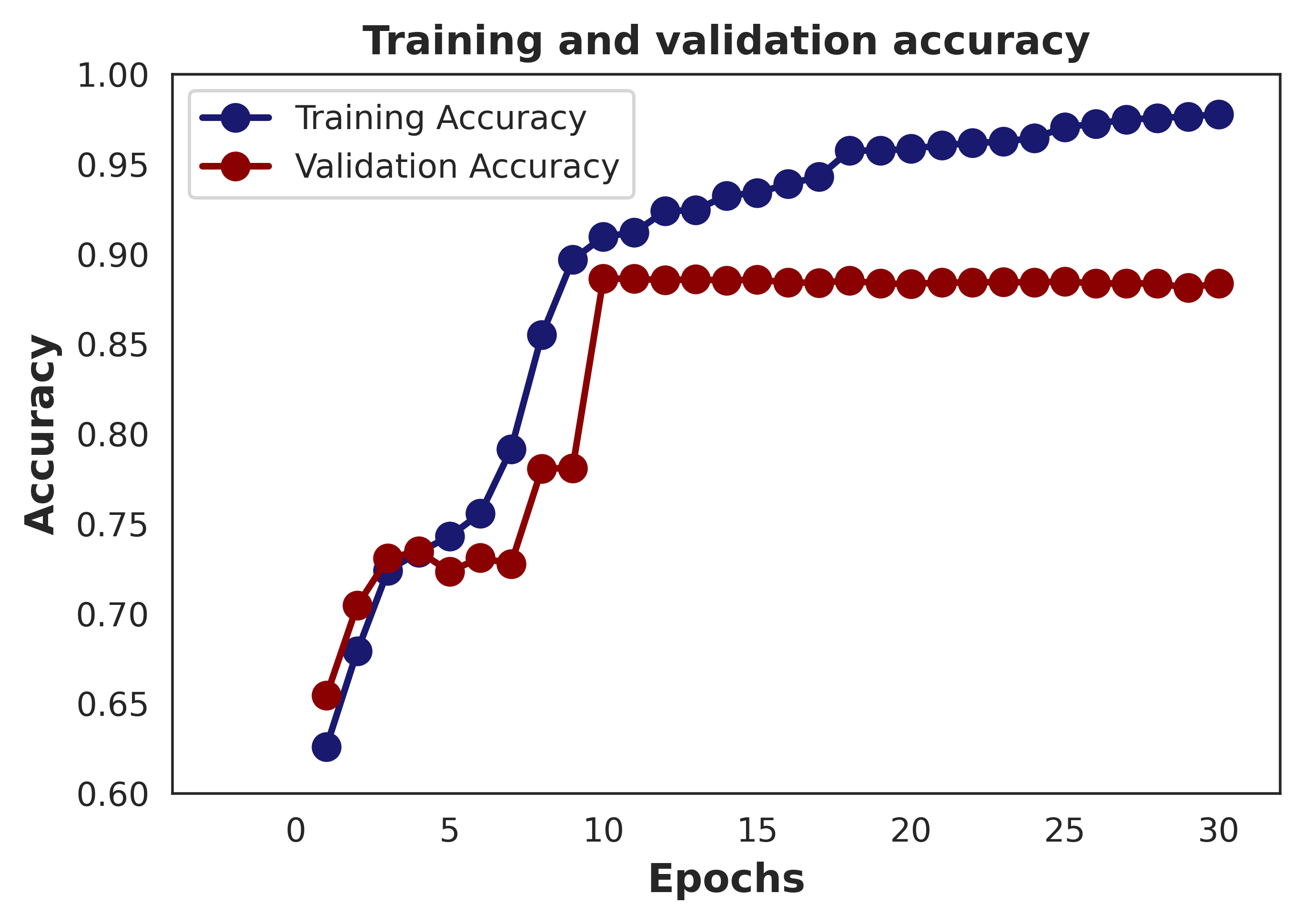}
\caption{Accuracy vs. number of epochs}
\end{subfigure}
\begin{subfigure}[h!]{0.49\textwidth}
\includegraphics[height=4.5cm, width=\textwidth]{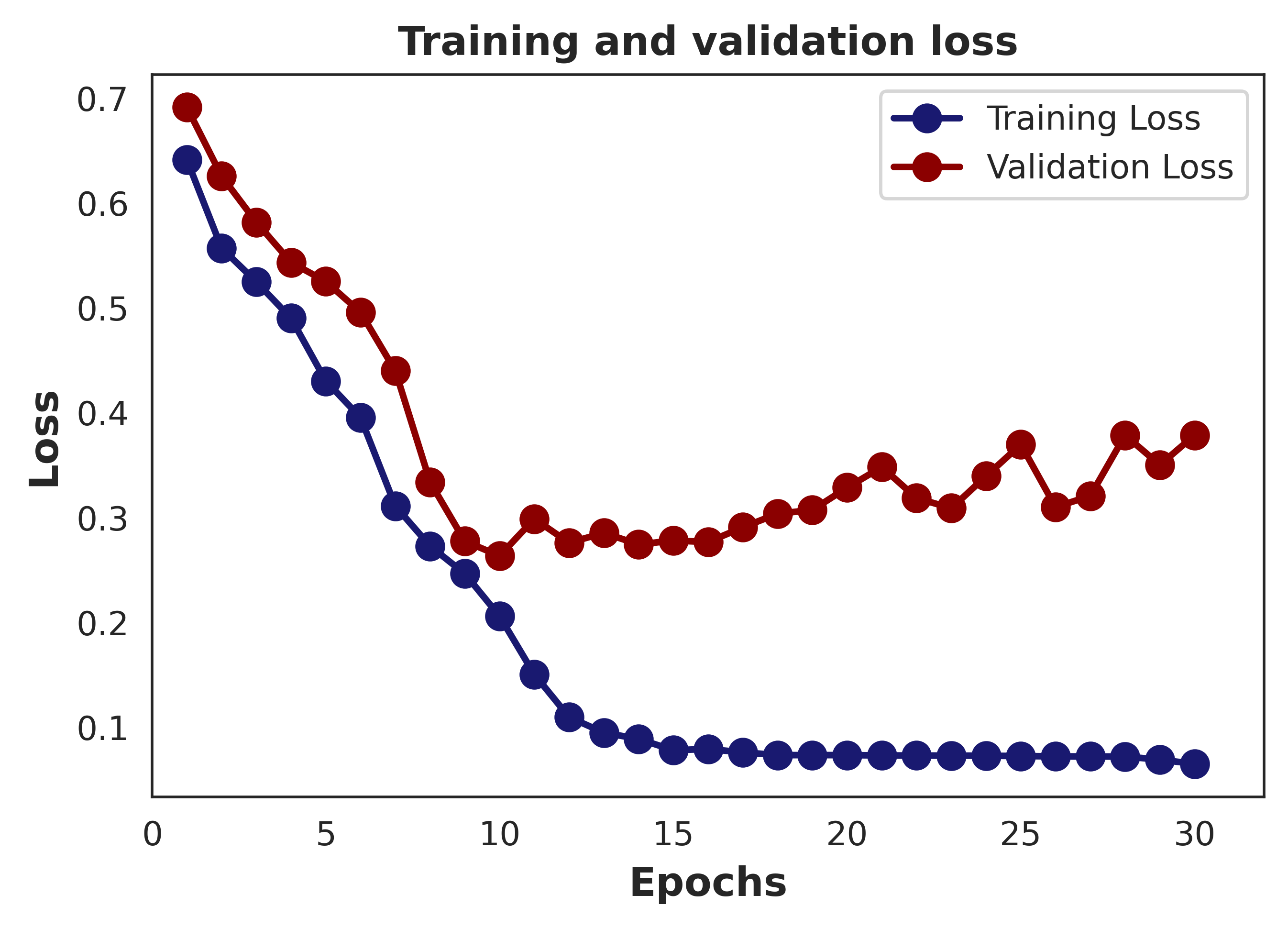}
\caption{Loss vs. number of epochs}
\end{subfigure}
\caption{Accuracy and loss variation during the training/validation phases.}
\label{acc-loss}
\end{figure}
Fig.~\ref{acc-loss} shows the accuracy and loss variation over the number of epochs for training and validation data. Initially, the validation accuracy is greater than the training accuracy, but after ten epochs, training accuracy keeps increasing while the validation accuracy stands at a flat line. Similarly, the validation loss also keeps decreasing with the training loss. 
At epoch ten, the validation accuracy reaches its maximum value of $88\%$. That is why we choose $10$ as the optimum value of the epoch number.

Furthermore, the performance of the proposed model is assessed on the test set by using various evaluation metrics such as accuracy, precision, recall and f1-score are noteworthy. On the other hand, the precision-recall (PR) curve and receiver operating characteristics (ROC) curve taken as the graphical evaluation metric. Apart from that, to analyze the effectiveness of the model, we compared its performance with different machine learning (ML) models. Logistic regression (LR), decision tree (DT), random forest (RF), Naive Bayes (NB) and support vector machine (SVM)  \cite{kowsari2019text} taken as the baseline models for comparison.
Term frequency-inverse document frequency \cite{sharif2019sentiment} (TF-IDF) feature representation tech-nique employed to train the ML algorithms on the same corpus. Finally, the similar test set used to evaluate all the models. The performance comparison of the models listed in Table~\ref{performance}.
\begin{table}[h!]
\begin{center}
\caption{Performance comparison with the ML models.}
\begin{tabular}{|C{3cm} | C{1.5cm}| C{1.5cm}|C{1.5cm}|C{1.5cm}|}
\hline
   Approach & Accuracy & Precision & Recall & F1-score \\
\hline
     Logistic Regression & $85.0$ & $82.0$ & $93.7$ & $87.5$  \\
\hline
    Decision Tree & $81.9$ & $83.5$ & $84.4$ & $83.9$ \\
\hline
    Random Forest & $84.7$ & $87.6$ & $84.8$ & $86.1$\\
\hline
 Naive Bayes & $89.5$ & $91.0$ & $90.1$ & $90.5$  \\
 \hline 
 Support Vector Machine & $88.3$ & $87.0$ & $93.0$ & $89.9$  \\
  \hline 
 Proposed (BiLSTM) & $\mathbf{91.35}$ & $93.1$ & $91.33$ & $\mathbf{ 92.21}$  \\
 \hline 
\end{tabular}
\label{performance}
\end{center}
\end{table}

\begin{figure}[b!]
       \centering
\begin{subfigure}[h!]{0.49\textwidth}
\includegraphics[height =4.5cm,width=\textwidth]{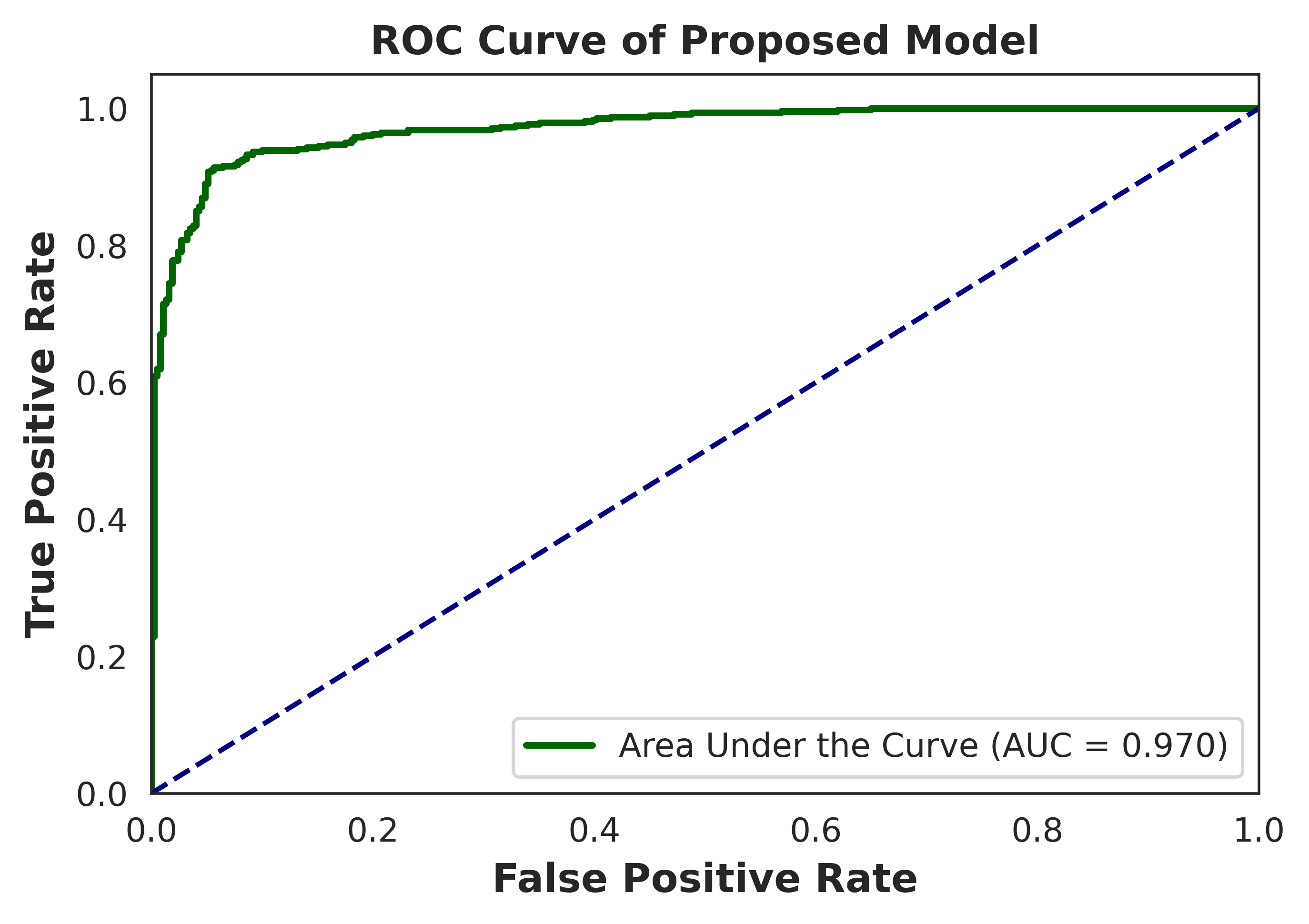}
\end{subfigure}
\begin{subfigure}[h!]{0.49\textwidth}
\includegraphics[height=4.5cm, width=\textwidth]{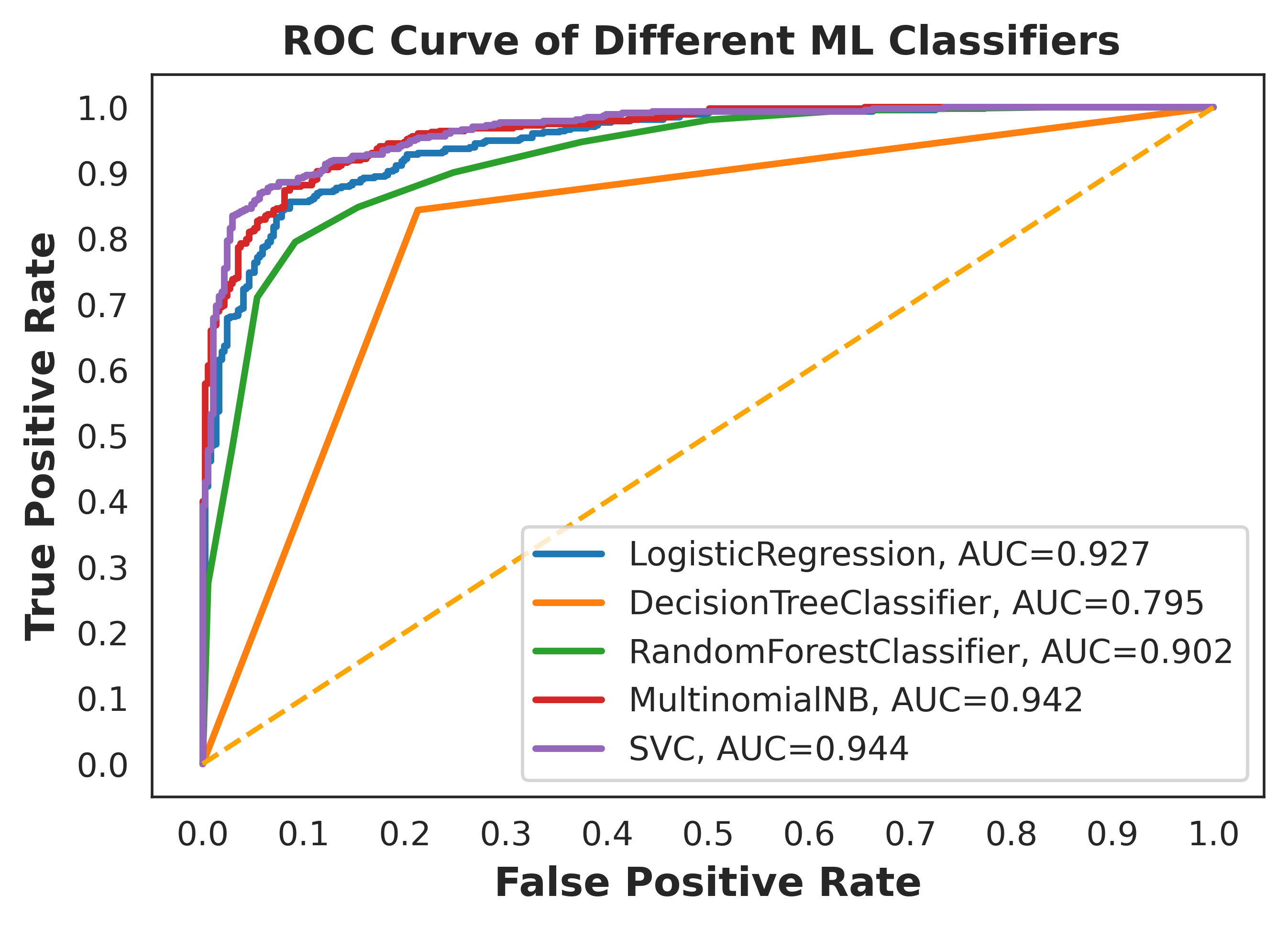}
\end{subfigure}
\caption{AUC comparison of the proposed and baseline model.}
\label{roc}
\end{figure}

From the table, it observed that SVM and NB provide pretty good accuracy of $88.3\%$ and $89.5\%$ respectively. Besides, they also give acceptable f1-scores of $89.9\%$ and $90.5\%$ respectively. Among the models, decision tree performs poorly compared to others. The proposed model outperforms all the ML models by achieving maximum accuracy of $91.35\%$ and f1-score of $92.21\%$.

\begin{figure}[t!]
       \centering
\begin{subfigure}[h!]{0.49\textwidth}
\includegraphics[height =4.5cm,width=\textwidth]{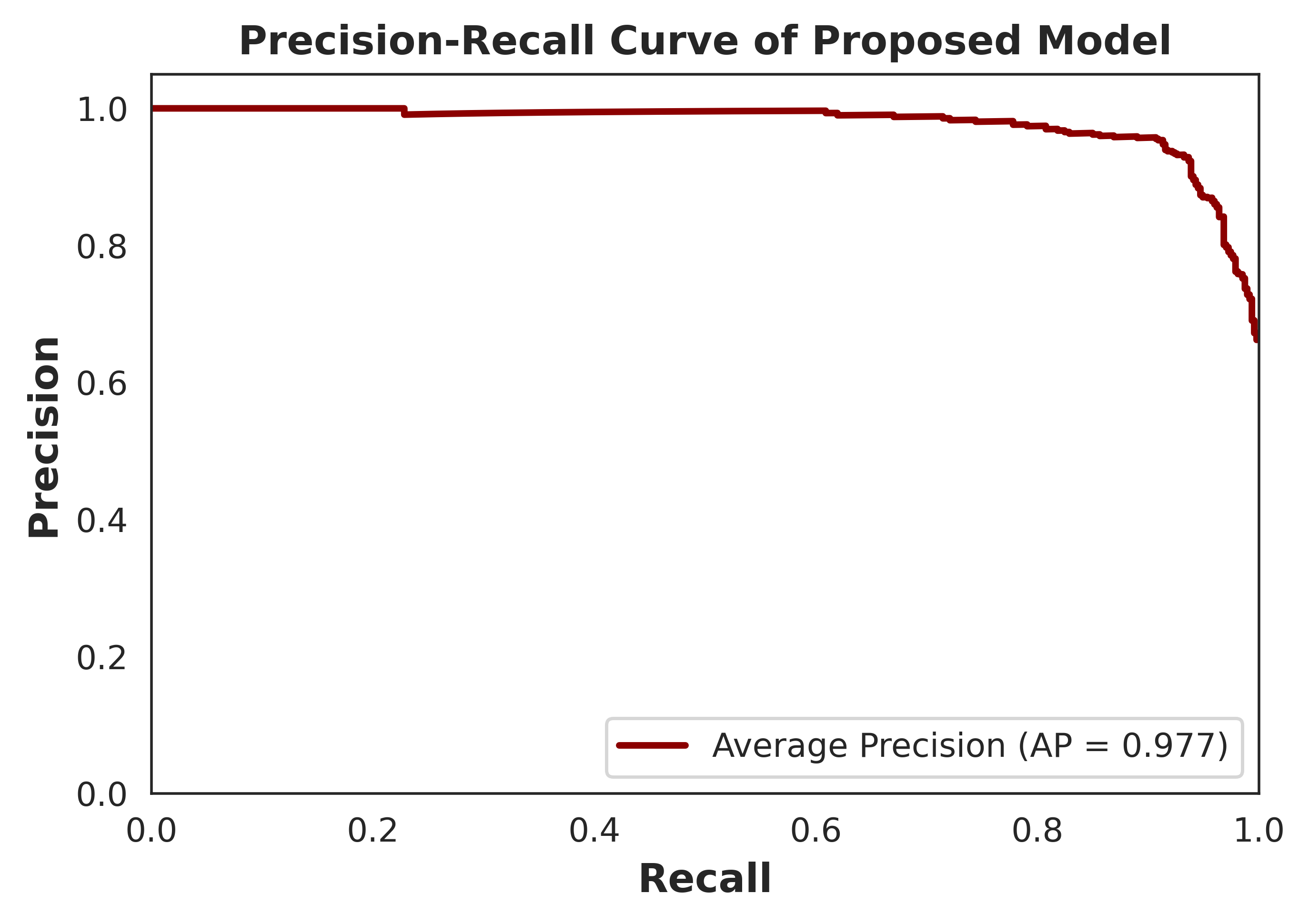}
\end{subfigure}
\begin{subfigure}[h!]{0.49\textwidth}
\includegraphics[height=4.5cm, width=\textwidth]{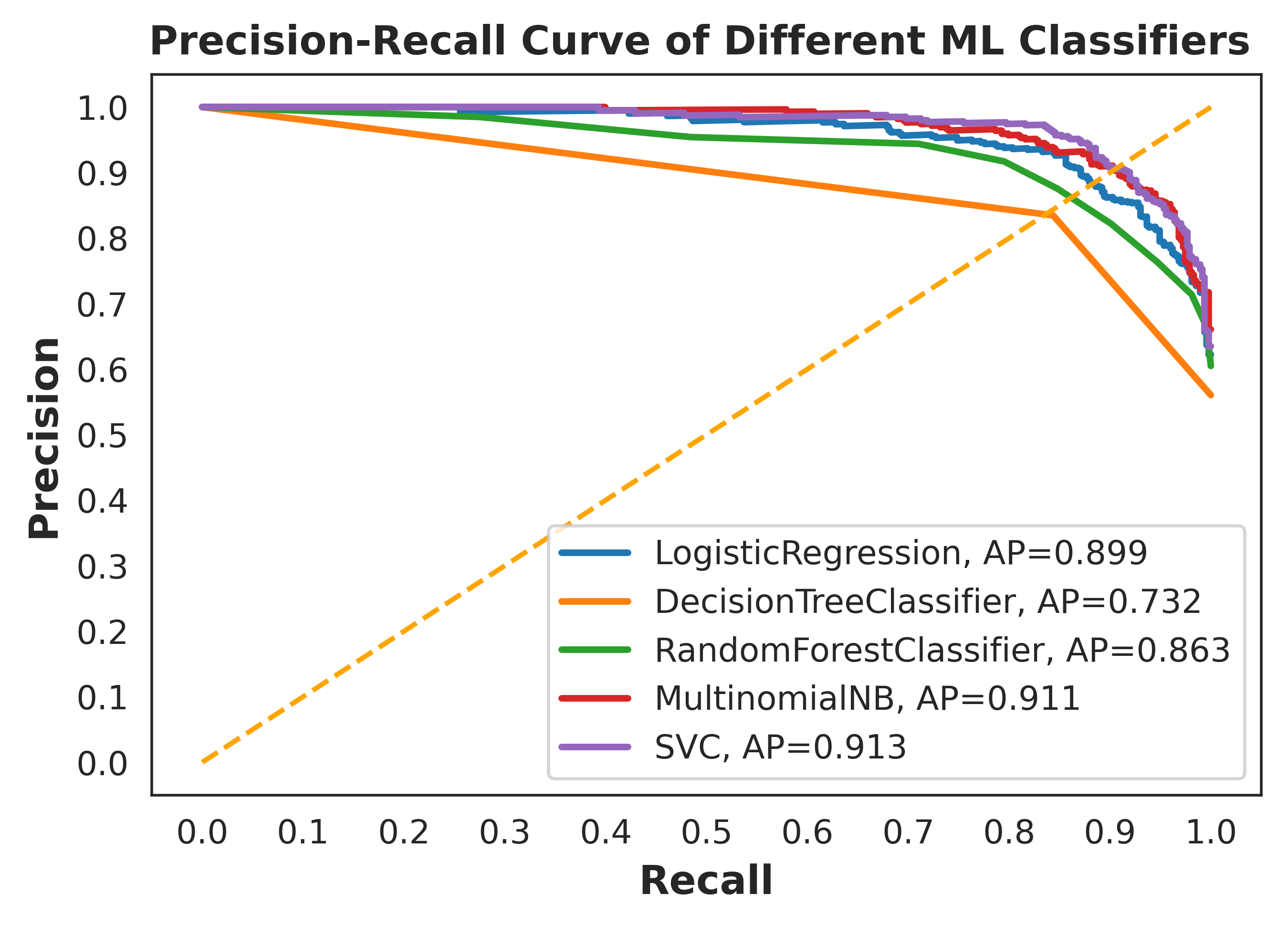}
\end{subfigure}
\caption{AP comparison of the proposed and baseline model.}
\label{pr}
\end{figure}

Fig.~\ref{roc} presents the ROC curve analysis of the proposed and baseline ML models. Among the ML models, SVM provides the highest area under the curve (AUC) value of approximately $95\%$ and DT provides the lowest AUC value of approximately $80\%$. On the contrary, the proposed BiLSTM model outperforms all the baseline models by acquiring AUC value of $97\%$. PR curve analysis of the models depicted in Fig.~\ref{pr}. The proposed model outdoes all the baseline models by achieving $97\%$ average precision (AP) score. Therefore, after analyzing the results, it is needless to say that the proposed model acquired the highest value for all the evaluation parameters.

To verify the assessment of the proposed method, we compared it with other techniques accomplished on similar tasks. We applied existing techniques on our developed corpus and compared the outcomes with the proposed approach. Table~\ref{comparison} shows the comparative analysis of the adopted techniques and their obtained accuracy on BRRC. The result indicates that the proposed approach performs better than other techniques. To sum up, it revealed that the proposed model shows outstanding result compared to the baseline models as well as the existing techniques.

\begin{table}[h!]
\begin{center}
\caption{Performance comparison with existing techniques.}
\begin{tabular}{|C{5cm} | C{5cm}|}
\hline
   Techniques & Accuracy (\%) on BRRC  \\
\hline
     CNN \cite{rahman2018aspect} & $87.7$   \\
\hline
    SVM + TF-IDF \cite{yu2017identifying}  & $88.3$  \\
 \hline 
 MNB + TF-IDF \cite{sharif2019sentiment}   & $89.5$   \\
  \hline 
 \textbf{Proposed (BiLSTM)} &  $\mathbf{91.35}$   \\
 \hline 
\end{tabular}
\label{comparison}
\end{center}
\end{table}

\section{Conclusion}
In this paper, we presented a deep learning-based scheme to analyze the sentiment on Bengali restaurant reviews. Word2vec embedding technique is used to consider the semantic meaning of the Bengali reviews. BiLSTM network tuned to find out the optimal hyperparameter combination. A corpus of 8435 Bengali restaurant reviews is developed to evaluate the performance of the proposed system. The outcome of the experimentation exhibits that the proposed system outperforms the baseline ML algorithms and previous techniques on a holdout dataset. Though our approach acquires satisfactory results compared to other works, some improv-ements are still required to take this system in production level. Thus, in future, we will try to add reviews with more classes and conjoin the aspect of the reviews as well.

%
%
\bibliographystyle{splncs04}
\bibliography{ref}

\end{document}